\documentclass[letterpaper, 10 pt, conference]{ieeeconf}  % Comment this line out if you need a4paper

\IEEEoverridecommandlockouts                              % This command is only needed if 
\usepackage{color,xcolor}
\usepackage{epsfig}
\usepackage{graphicx}

\usepackage{adjustbox}
\usepackage{array}
\usepackage{booktabs}
\usepackage{colortbl}
\usepackage{color}
\usepackage{float,wrapfig}
\usepackage{hhline}
\usepackage{multirow}
\usepackage{subcaption} % issues a warning with CVPR/ICCV format

\usepackage{makecell}
\usepackage{amsmath,amsfonts,amssymb}
\usepackage{bm}
\usepackage{nicefrac}
\usepackage{microtype}

\usepackage{changepage}
\usepackage{extramarks}
\usepackage{fancyhdr}
\usepackage{lastpage}
\usepackage{setspace}
\usepackage{soul}
\usepackage{xspace}

\usepackage{url}

\usepackage{algorithm, algorithmic}
\usepackage{enumerate}
\usepackage{todonotes} % conflict with CVPR/ICCV format
\newcolumntype{L}[1]{>{\raggedright\let\newline\\\arraybackslash\hspace{0pt}}m{#1}}
\newcolumntype{C}[1]{>{\centering\let\newline\\\arraybackslash\hspace{0pt}}m{#1}}
\newcolumntype{R}[1]{>{\raggedleft\let\newline\\\arraybackslash\hspace{0pt}}m{#1}}

\newcommand{\fig}[1]{Fig.~\ref{#1}}

\newcommand{\tbl}[1]{Table~\ref{#1}}

\newcommand{\ignorethis}[1]{}

\makeatletter
\DeclareRobustCommand\onedot{\futurelet\@let@token\@onedot}
\def\@onedot{\ifx\@let@token.\else.\null\fi\xspace}

\def\eg{e.g\onedot}

\def\wrt{w.r.t\onedot} 
\def\etal{et al\onedot}
\makeatother

\definecolor{MyDarkBlue}{rgb}{0,0.08,1}
\definecolor{MyDarkGreen}{rgb}{0.02,0.6,0.02}
\definecolor{MyDarkRed}{rgb}{0.8,0.02,0.02}
\definecolor{MyDarkOrange}{rgb}{0.40,0.2,0.02}
\definecolor{MyPurple}{RGB}{111,0,255}
\definecolor{MyRed}{rgb}{1.0,0.0,0.0}
\definecolor{MyGold}{rgb}{0.75,0.6,0.12}
\definecolor{MyDarkgray}{rgb}{0.66, 0.66, 0.66}

\newcommand{\yh}[1]{\textcolor{MyDarkBlue}{[Yuanming: #1]}}

\newcommand{\Name}{ChainQueen\xspace}
\usepackage{CJKutf8}

\newcommand{\xx}{\mathbf{x}}

\newcommand{\vv}{\mathbf{v}}

\newcommand{\WW}{\mathbf{W}}

\newcommand{\CC}{\mathbf{C}}

\newcommand{\pp}{\mathbf{p}}

\renewcommand{\gg}{\mathbf{g}}
\newcommand{\GG}{\mathbf{G}}

\newcommand{\FF}{\mathbf{F}}

\newcommand{\PP}{\mathbf{P}}

\newcommand{\zz}{\mathbf{z}}

\renewcommand{\tt}{\mathbf{t}}
\newcommand{\II}{\mathbf{I}}
\newcommand{\sig}{\boldsymbol\sigma}

\newcommand{\nn}{\mathbf{n}}

\newcommand{\dt}{\Delta{}t}

\renewcommand{\AA}{\mathbf{A}}

\newcommand\restr[2]{{\left.\kern-\nulldelimiterspace{}#1\right|_{#2}}}

\title{\LARGE \bf
\Name: A Real-Time Differentiable Physical Simulator\\for Soft Robotics}

\author{Yuanming Hu, Jiancheng Liu$^*$, Andrew Spielberg$^*$,\\Joshua B. Tenenbaum, William T. Freeman, Jiajun Wu, Daniela Rus, Wojciech Matusik$^{1,2}$
\thanks{$^{1}$Y. Hu, A. Spielberg, J. B. Tenenbaum, W. T. Freeman, J. Wu, D. Rus, and W. Matusik are with Computer Science and Artificial Intelligence Laboratory, Massachusetts Institute of Technology, Cambridge, MA, USA}
\thanks{$^{2}$J. Liu is with Institute for Interdisciplinary Information Science, Tsinghua University, Beijing, China}
\thanks{$^{*}$ Equally contributed.}
}

\begin{document}

\maketitle
\thispagestyle{empty}
\pagestyle{empty}

\newcommand{\D}[1]{\frac{\partial L}{\partial {{#1}}}}
\newcommand{\pd}[2]{\frac{\partial{#1}}{\partial{#2}}}
\newcommand{\chain}[2]{\D{#1} \pd{#1}{#2}}
\newcommand{\dep}{\leftarrow}
\newcommand{\rdep}{\rightarrow}
\newcommand{\ttau}{\boldsymbol{\tau}}
\newcommand{\Dinv}{\frac{4}{\Delta x^2}}
\newcommand{\ePpn}{\PP_p^n & = & \PP_p^n(\FF_p^n)+\FF_p\sig_{pa}^n}
\newcommand{\emin}{m_i^{n} & = & \sum_p N(\xx_i-\xx_p^n)m_p}
\newcommand{\eppin}{\pp_i^{n} & = & \sum_p N(\xx_i- \xx_p^n) \left[m_p\vv_p^n+\left(-\Dinv\Delta t V_p^0\PP_p^n\FF_p^{nT} + m_p \CC_p^n\right)(\xx_i-\xx_p^n)\right]}
\newcommand{\evvin}{\vv_i^{n} & = & \frac{1}{m_i^n}\pp_i^n}
\newcommand{\evvpnn}{\vv_p^{n+1} & = & \sum_i N(\xx_i- \xx_p^n)\vv_i^n}
\newcommand{\eCCpnn}{\CC_p^{n+1} & = &\frac{4}{\Delta x^2} \sum_i N(\xx_i- \xx_p^n)\vv_i^n(\xx_i-\xx_p^n)^T}
\newcommand{\eFFpnn}{\FF_p^{n+1} & = & (\II + \Delta t \CC_p^{n+1})\FF_p^n}
\newcommand{\exxpnn}{\xx_p^{n+1} & = & \xx_p^n + \Delta t \vv_p^{n+1}}

%%%%%%%%%%%%%%%%%%%%%%%%%%%%%%%%%%%%%%%%%%%%%%%%%%%%%%%%%%%%%%%%%%%%%%%%%%%%%%%%
\begin{abstract}

Physical simulators have been widely used in robot planning and control. Among them, differentiable simulators are particularly favored, as they can be incorporated into gradient-based optimization algorithms that are efficient in solving inverse problems such as optimal control and motion planning. 
Simulating deformable objects is, however, more challenging compared to rigid body dynamics. The underlying physical laws of deformable objects are more complex, and the resulting systems have orders of magnitude more degrees of freedom and therefore they are significantly more computationally expensive to simulate. Computing gradients with respect to physical design or controller parameters is typically even more computationally challenging. In this paper, we propose a real-time, differentiable hybrid Lagrangian-Eulerian physical simulator for deformable objects, \Name, based on the Moving Least Squares Material Point Method (MLS-MPM). MLS-MPM can simulate deformable objects including contact 
and can be seamlessly incorporated into inference, control and co-design systems.
We demonstrate that our simulator achieves high precision in both forward simulation and backward gradient computation. We have successfully employed it in a diverse set of control tasks for soft robots, including problems with nearly 3,000 decision variables.     
\end{abstract}

\vspace{-5pt}
\section{Introduction}
\vspace{-5pt}

Robot planning and control algorithms often rely on physical simulators for prediction and optimization~\cite{Todorov2012MuJoCo,Erez2015Simulation}. In particular, differentiable physical simulators enable the use of gradient-based optimizers, significantly improving control efficiency and precision. 
Motivated by this, there has been extensive research on differentiable rigid body simulators, using approximate~\cite{chang2016compositional, mrowca2018flexible} and exact~\cite{degrave2016differentiable, drake, frigerio:2016:robcogen} methods.

Significant challenges remain for deformable objects. First, simulating the motion of deformable objects is slow, because they have much higher degrees of freedom (DoFs). 
Second, contact detection and resolution is challenging for deformable objects, due to their changing geometries and potential self-collisions. Third, closed-form and efficient computation of gradients is challenging in the presence of contact. As a consequence, current simulation methods for soft objects cannot be effectively used for solving inverse problems such as optimal control and motion planning. 

In this paper, we introduce a real-time, differentiable physical simulator for deformable objects, building upon the Moving Least Squares Material Point Method (MLS-MPM)~\cite{hu2018moving}. We name our simulator \Name\footnote{Or
\begin{CJK}{UTF8}{gbsn}
乾坤
\end{CJK}
, literally ``everything between the sky and the earth.''}. The Material Point Method (MPM) is a hybrid Lagrangian-Eulerian method that uses both particles and grid nodes for simulation~\cite{sulsky1995application}. MLS-MPM accelerates and simplifies traditional MPM using a moving least squares force discretization. In \Name, we introduce the first fully differentiable MLS-MPM simulator with respect to both state and model parameters, with both forward simulation and back-propagation running efficiently on GPUs. We demonstrate the ability to efficiently calculate gradients with respect to the entire simulation. This enables many novel applications for soft robotics including optimization-based closed-loop controller design, trajectory optimization, and co-design of robot geometry, materials, and control.

As a particle-grid-based hybrid simulator, MPM simulates objects of various states, such as liquid (\eg, water), granular materials (\eg, sand), and elastoplastic materials (\eg, snow and human tissue). 
\Name focuses on elastic materials for soft robotics. It is fully differentiable and $4-9\times$ faster than the current state-of-the-art. 
Numerical and experimental validation suggest that \Name
achieves high precision in both forward simulation and backward gradient computation. 

\Name's differentiability allows it to support gradient-based optimization for control and system identification. By performing gradient descent on controller parameters, our simulator is capable of solving these inverse problems on a diverse set of complex tasks, such as 
optimizing a 3D soft walker controller given an objective.
Similarly, gradient descent on physical design parameters,
enables inference of physical properties (e.g. mass, density and Young's modulus) of objects and optimizing design for a desired task.

In addition to benchmarking \Name's performance and demonstrating its capabilities on a diverse set of inverse problems, we have interfaced our simulator with high-level python scripts to make \Name user-friendly. Users at all levels will be able to develop their own soft robotics systems using our simulator, without the need to understand its low-level details. We will open-source our code and data and we hope they can benefit the robotics community.
\section{Related Work}

\subsection{Material Point Method}
The material point method has been extensively developed from both a solid mechanics~\cite{sulsky1995application} and computer graphics~\cite{jiang2016material} perspective. As a hybrid Eulerian-Langrangian method, MPM has demonstrated its versatility in simulating snow~\cite{stomakhin2013material, gaume2018dynamic}, sand~\cite{klar2016drucker, daviet2016semi}, non-Newtonion fluids~\cite{ram2015material}, cloth~\cite{jiang2017anisotropic, guo2018material}, solid-fluid coupling~\cite{Gao2018sedimentfluid,Tampubolon2017multispecies}, rigid body coupling, and cutting~\cite{hu2018moving}. \cite{Gao2017adaptivempm} also proposed an adaptive MPM scheme to concentrate computation resources in the regions of interest.

There are many benefits of using MPM for soft robotics. First, MPM is a well-founded and physically-accurate discretization method and can be derived through the weak form of conservation laws. Such a physically-based approach makes it easier to match simulation with real-world experiments. Second, MPM is friendly to parallelization on modern hardware architectures. 
Closely related to our work is a high-performance GPU implementation~\cite{gao2018gpu} by Gao \etal, from which we borrow many useful optimization practices. Though efficient when solving forward simulation, their simulator is not differentiable, making it inefficient for inverse problems in robotics and learning. Third, MPM naturally handles large deformation and (self-)collision, which are common in soft robotics, but often not modeled in, \eg, mesh-based approaches due to computational expense. Finally, the continuum dynamics (including soft object collision) are governed by the smooth (and differentiable) potential energy, making the whole system differentiable.

Our simulator, \Name, is fully differentiable and the first simulator that applies MPM to soft robotics. 

\subsection{Differentiable Simulation and Control}

Recently, there has been an increasing interest in building differentiable simulators for planning and control. For rigid bodies,
\cite{Battaglia2016Interaction}, 
\cite{chang2016compositional} and \cite{mrowca2018flexible} proposed to approximate object interaction with neural nets; later, 
\cite{Sanchez-Gonzalez2018Graph} explored their usage in control. 
Approximate analytic differentiable rigid body simulators have also been proposed~\cite{degrave2016differentiable,de2018modular}. Such systems have been deployed for manipulation and planning ~\cite{toussaint2018differentiable}.

Differentiable simulators for deformable objects have been less studied. Recently, 
\cite{schenck2018spnets} proposed SPNets for differentiable simulation of position-based fluids~\cite{macklin2013position}. The particle interactions are coded as neural network operations and differentiability is achieved via automatic differentiation in PyTorch. A hierarchical particle-based object representation using neural networks is also proposed in \cite{mrowca2018flexible}. Instead of approximating physics using neural networks, \Name differentiates MLS-MPM, a well physically founded discretization scheme derived from continuum mechanics. In summary, our simulator can be used for a more diverse set of objects; it is more physically plausible, and runs faster. 

\begin{figure*}
    \centering
    \includegraphics[width=1.0\linewidth]{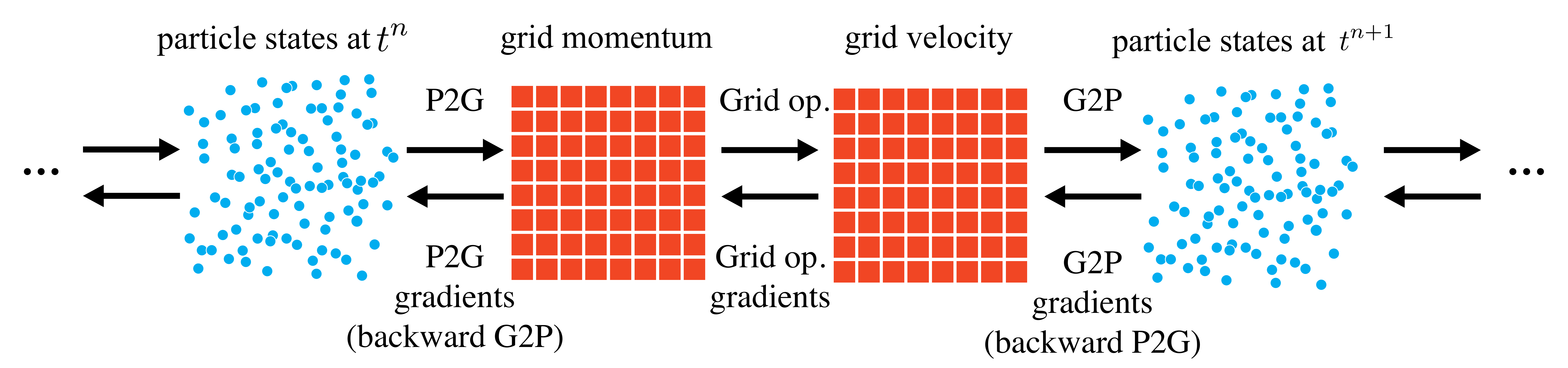}
    \caption{One time step of MLS-MPM. Top arrows are for forward simulation and bottom ones are for back propagation. A controller is embedded in the P2G process to generate actuation given particle configurations.}
    \label{fig:mpm}
    \vspace{-15pt}
\end{figure*}
\begin{table}[t]
  \centering\small
  \setlength{\tabcolsep}{4pt}
\caption{List of notations for MLS-MPM.}
\begin{tabular}{cccl}
\toprule
Symbol & Type  & Affiliation & Meaning \\ 
\midrule
$\Delta t$ & scalar &   & time step size \\
$\Delta x$ & scalar &   & grid cell size \\
$\xx_p$ & vector & particle & position \\
$V_p^0$ & scalar  & particle & initial volume\\
$\vv_p$ & vector & particle & velocity \\
$\CC_p$ & matrix & particle & affine velocity field ~\cite{jiang:2015:apic} \\
$\PP_p$ & matrix & particle &  
PK1 stress ($\partial \psi_p/\partial \FF_p$)\\
$\sig_{pa}$ & matrix & particle &  
actuation Cauchy stress \\
$\AA_{p}$ & matrix & particle &  
actuation stress (material space) \\
$\FF_p$ & matrix & particle & deformation gradient\\
$\xx_i$ & vector & node  & position\\
$m_i$ & scalar  & node & mass\\
$\vv_i$ & vector & node & velocity\\
$\pp_i$ & vector & node & momentum, i.e. $m_i\vv_i$\\
$N$ & scalar &  & quadratic B-spline function\\
\bottomrule
\end{tabular}%
\label{table:notations}
\vspace{-15pt}
\end{table}

\section{Forward simulation and back-propagation}
We use the moving least squares material point method (MLS-MPM)~\cite{hu2018moving}
to discretize continuum mechanics, which is governed by the following two equations:
\begin{align}
  \rho\frac{D\vv}{Dt} = \nabla \cdot \sig + \rho \gg \ \ \  &\text{(momentum conservation)}, \label{eqn:conservation-momentum}\\
  \frac{D\rho}{Dt} + \rho \nabla \cdot \vv = 0  \ \ \  &\text{(mass conservation)}.
\end{align}

We briefly cover the basics of MLS-MPM and readers are referred to \cite{jiang2016material} and \cite{hu2018moving} for a comprehensive introduction of MPM and MLS-MPM, respectively. The material point method is a hybrid Eulerian-Lagrangian method, where both particles and grid nodes are used. Simulation state information is transferred back-and-forth between these two representations. We summarize the notations we use in this paper in Table ~\ref{table:notations}. Subscripts are used to denote particle ($p$) and grid nodes ($i$), while superscripts ($n$, $n+1$) are used to distinguish quantities in different time steps.
The MLS-MPM simulation cycle has three steps:
\begin{enumerate}
\item \textbf{Particle-to-grid transfer (P2G).}
Particles transfer mass $m_p$, momentum $(m\vv)_p^n$, and stress-contributed impulse to their neighbouring grid nodes, using the Affine Particle-in-Cell method (APIC)~\cite{jiang:2015:apic} and moving least squares force discretization~\cite{hu2018moving}, weighted by a compact B-spline kernel $N$:
\small
\begin{eqnarray}
\emin,\\
\GG_p^n&=&-\Dinv\Delta t V_p^0\PP_p^n\FF_p^{nT} + m_p \CC_p^n,\\
\small\pp_i^{n} & = & \sum_p N(\xx_i- \xx_p^n) \left[m_p\vv_p^n+\GG_p^n(\xx_i-\xx_p^n)\right].
\end{eqnarray}
\normalsize
\vspace{-8pt}
\item \textbf{Grid operations.} Grid momentum is normalized into grid velocity after division by grid mass: \begin{eqnarray}
\evvin.
\end{eqnarray}
Note that neighbouring particles interact with each other through their shared grid nodes, and collisions are handled automatically. Here we omit boundary conditions and gravity for simplicity.
\item \textbf{Grid-to-particle transfer (G2P).} Particles gather updated velocity $\vv_p^{n+1}$, local velocity field gradients $\CC_p^{n+1}$ and position $\xx_p^{n+1}$. The constitutive model properties (e.g. deformation gradients $\FF_p^{n+1}$) are updated.

\small
\begin{eqnarray}
\evvpnn, \\
\eCCpnn, \\
\eFFpnn, \\
\exxpnn.
\end{eqnarray}
\normalsize

\end{enumerate}

For soft robotics, we additionally introduce an actuation model. Inspired by actuators such as \cite{hara2004artificial}, we designed an actuation model that expands or stretches particle $p$ via an additional Cauchy stress $\AA_p=\FF_p\sig_{pa}^{}\FF_p^T$, with $\sig_{pa}=\text{Diag}(a_x, a_y, a_z)$ -- the stress in the material space. This framework supports the use of other differentiable actuation models including pneumatic, hydraulic, and cable-driven actuators. \fig{fig:mpm} illustrates forward simulation and back propagation.

MLS-MPM is naturally differentiable. Though the forward direction has been extensively used in computer graphics, the backward direction (differentiation or back-propagation) is largely unexplored.

\begin{figure*}[t]
    \centering
    \includegraphics[width=\textwidth]{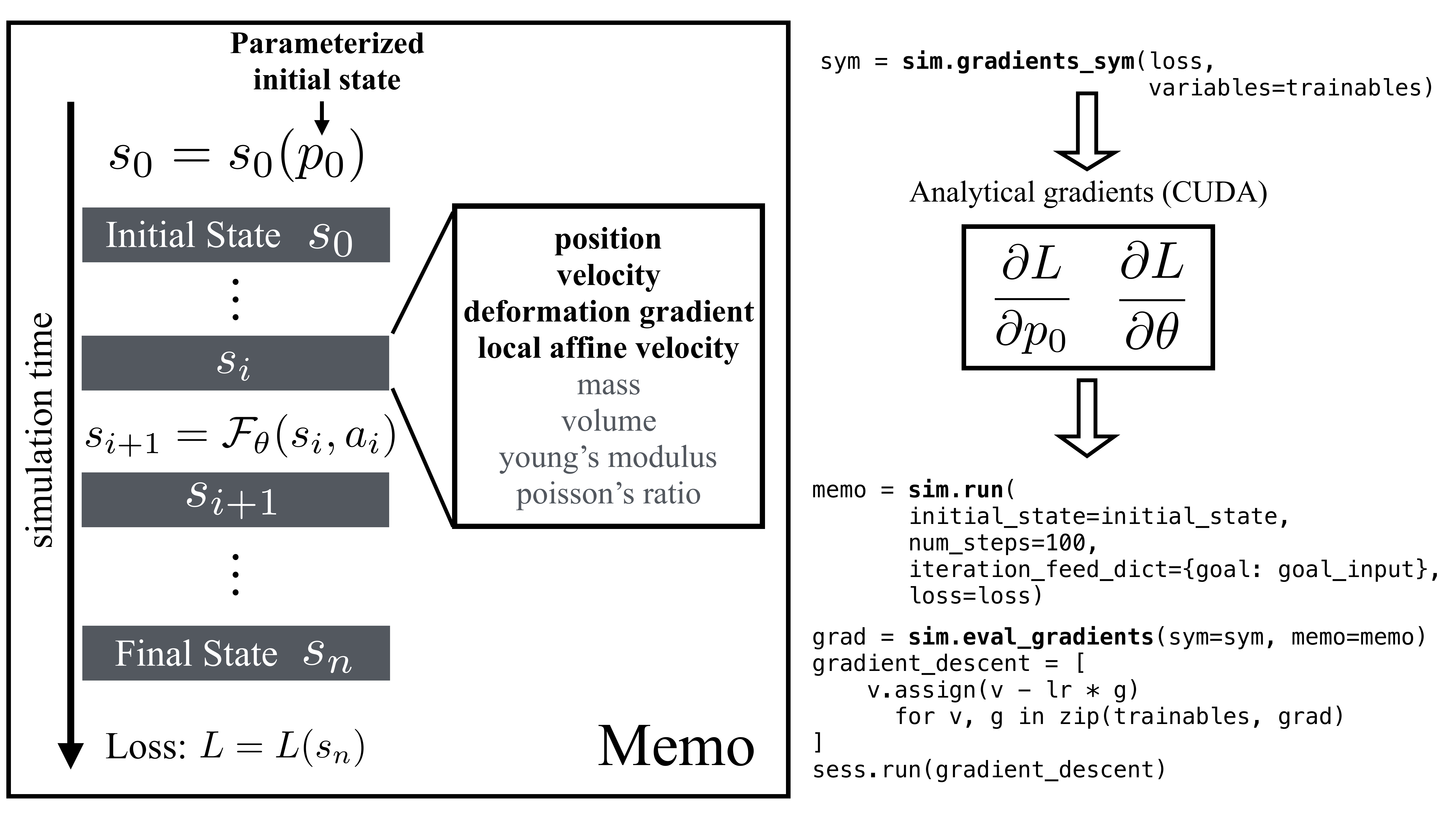}
    \caption{Left: A ``memo" object consists all information of a single simulation execution, including all the time step state information (position, velocity, deformation gradients etc.), and parameters for the initial state $p_0$, policy parameter $\theta$. Right: Code samples to get the symbolic differentiation (top) and memo, evaluate gradients out of the memo and symbolic differentiation, and finally use them for gradient descent (bottom).}
    \label{fig:differentiation}
    \vspace{-15pt}
\end{figure*}

Based on the gradients we have derived analytically, we have designed a high-performance implementation that resembles the traditional forward MPM cycle: backward P2G (scatter particle gradients to grid), grid operations, and backward G2P (gather grid gradients to particles).
\footnote{Please see the supplemental document 
for the gradient derivations.}
Gradients of state at the end of a time step with respect to states at the starting of the time step can be computed using the chain rule. With the single-step gradients computed, applying the chain rule at a higher level from the final state all-the-way to the initial state yields gradients of the final state with respect to the initial state, as well as to the controller parameters that are used in each state. We cache all the simulation states in memory, using a ``memo'' object. Though the underlying differentiation is complicated, we have designed a simple high-level TensorFlow interface on which end-users can build their applications (\fig{fig:differentiation}).

Our high-performance implementation\footnote{Based the \textbf{Taichi}~\cite{hu2018taichi} open source computer graphics library.} takes advantage of the computational power of modern GPUs through CUDA. We also implemented a reference implementation in TensorFlow. Note that programming physical simulation as a ``computation graph'' using high-level frameworks such as TensorFlow is less inefficient. In fact, when all the overheads are gone, our optimized CUDA solver is $132\times$ faster than the TensorFlow reference version. This is because TensorFlow is optimized towards deep learning applications where data granularity is much larger and memory access pattern is much more regular than physical simulation, and limited CPU-GPU bandwidth. In contrast, our CUDA implementation is tailored for MLS-MPM and explicitly optimized for parallelism and locality, thus delivering high-performance.

\section{Evaluation}

\begin{table}[t]
  {\centering \small
\caption{Performance comparisons on a NVIDIA GTX 1080 Ti GPU. \textbf{F} stands for forward simulation and
\textbf{B} stands for backward differentiation. \textbf{TF} indicates the TensorFlow implementation. When benchmarking our simulator with CUDA we use the C++ instead of python interface to avoid the extra overhead due to the TensorFlow runtime library.}
\begin{tabular}{lccc}
\toprule
Approach                        & Impl.& \# Particles  & Time per Frame  \\ 
\midrule
Flex (3D)           & CUDA & 8,024   & 3.5 ms (286 FPS)  \\
Ours (3D, F)        & CUDA & 8,000   & 0.392 ms (\textbf{2,551} FPS) \\ 
Ours (3D, B)        & CUDA & 8,000   & 0.406 ms (\textbf{2,463} FPS) \\ 
\midrule
Flex (3D)           & CUDA & 61,238  & 6 ms (167 FPS)\\
Ours (3D, F)        & CUDA & 64,000  & 1.594 ms (\textbf{628} FPS)\\
Ours (3D, B)        & CUDA & 64,000  & 1.774 ms (\textbf{563} FPS) \\ \midrule
Ours (3D, F)        & CUDA & 512,000  & 10.501 ms (\textbf{92} FPS) \\
Ours (3D, B)        & CUDA & 512,000  & 11.594 ms (\textbf{86} FPS) \\ 
\midrule
\midrule
Ours (2D, F)        & TF   & 6,400 & 13.2 ms (76 FPS) \\
Ours (2D, B)        & TF   & 6,400 & 35.7 ms (28 FPS) \\
Ours (2D, F)        & CUDA & 6,400 & 0.10 ms (\textbf{10,000 FPS})\\ Ours (2D, B)        & CUDA & 6,400 & 0.14 ms (\textbf{7,162 FPS}) \\ 
\bottomrule
\end{tabular}%
\label{table:speed}}
\vspace{-15pt}
\ \\

%{\footnotesize
%\par
%}
\end{table}

In this section, we conduct a comprehensive study of the efficiency and accuracy of our system, in both 2D and 3D.

\subsection{Efficiency}
Instead of using complex geometries, a simple falling cube is used for performance benchmarking, to ensure easy analysis and reproducibility.
We benchmark the performance of our CUDA simulator against NVIDIA Flex~\cite{macklin2014unified}, a popular PBD physical simulator capable of simulating deformable objects.
Note that both PBD and MLS-MPM needs substepping iterations to ensure high stiffness. To ensure fair comparison, we set a Young's modulus, Poisson's ration and density so that visually \Name gives similar results to Flex. We used two steps per frame and four iterations per step in Flex. Note that setting exactly the same parameters is not possible since in PBD there is no explicitly defined physical quantity such as Young's modulus. 

We summarize the quantitative performance in \tbl{table:speed}. Our CUDA simulator provides higher speed than Flex, when the number of particles are the same. It is also worth noting that the TensorFlow implementation is much slower, due to excessive runtime overheads.

\subsection{Accuracy}
We design five test cases to evaluate the accuracy of both forward simulation and backward gradient evaluation:
\begin{enumerate}
    \item A1 (analytic, 3D, float32 precision): final position w.r.t. initial velocity (with collision). This case tests conservation of momentum, gradient accuracy and stability of back-propagation.
    \item A2 (analytic, 3D, float32 precision): same as A1 but with one collision to a friction-less wall. 
    \item B (numeric, 2D, float64 precision): colliding billiards. This case tests gradient accuracy and stability in more complex cases where analytic solutions do not exist. We used float64 precision for accurate finite difference results.
    \item C (numeric, 2D, float64 precision): finger controller. This case tests gradient accuracy of controller parameters, which are used repeatedly in the whole simulation process.
    \item D1 (experimental, pneumatic actuator, actuation) In order to evaluate our simulator's real-world accuracy, we compared the deformation of a physical actuator to a virtual one.  The physical actuator has four pneumatic chambers which can be inflated with an external pump, arranged in a cross-shape.  Inflating the individual chambers bends the actuator away from that chamber.  The actuator was cast using Smooth-On Dragon Skin 30.
    \item D2 (experimental, pneumatic actuator, bouncing) In a second test, we dropped the same actuator from a 15~cm height, and compared its dynamic motion to a simulation.
\end{enumerate}

\begin{figure}
    \centering
    \includegraphics[width=\linewidth]{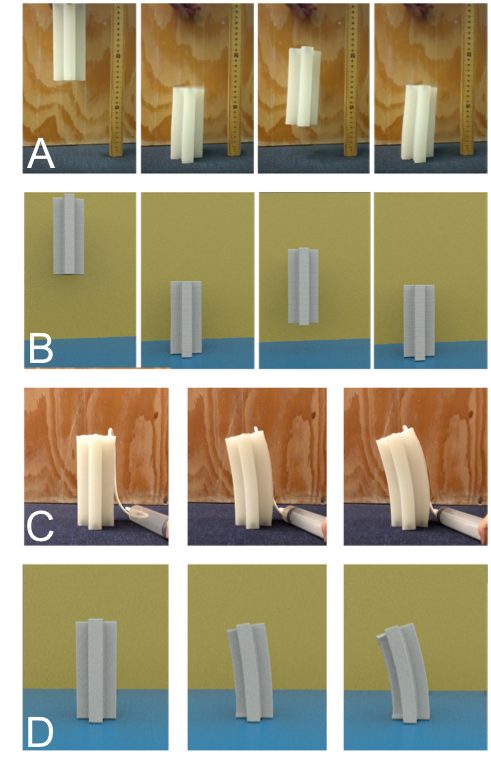}
    \vspace{-25pt}
    \caption{Experiments on the pneumatic leg. \textbf{Row (A, B)} Footage and simulator results of a bouncing experiment with the leg dropping at 15 cm. \textbf{Row (C, D)} Actuation test.}
    \vspace{-5pt}
    \label{fig:simreal}
\end{figure}

In 3D analytic test cases, where gradients \wrt initial velocity can be directly evaluated as in \tbl{table:accuracy}. For the experimental comparisons, the results are shown in \fig{fig:simreal}. In addition to our simulator's high performance and accuracy, it is worth noting that
that the gradients remain stable in the long term, within up to $1000$ time steps.

\begin{table}[t]
  \centering
  \small
  \setlength{\tabcolsep}{4pt}
  %\fontsize{.9em}{.6em}\selectfont
\caption{Relative error in simulation and gradient precision. Empty values are because of too short time for collision to happen.}
\begin{tabular}{lcccc}
\toprule
Case &   1 steps           & 10 steps & 100 steps & 1000 steps\\ 
\midrule
A1 & $9.80\times10^{-8}$ & $4.74\times10^{-8}$ & $1.15\times10^{-7}$ & $1.43\times10^{-5}$ \\
A2 & -  &  -    &  -     &  $2.69\times10^{-5}$   \\
B    &    $ $-   &    -     &  $2.39\times10^{-8}$  & $2.83\times10^{-8}$             \\
C    & $5.63\times10^{-6}$ &    $2.24\times10^{-7}$      &  $6.97\times10^{-7}$ & $1.76\times10^{-6}$ \\
\bottomrule
\end{tabular}%
\label{table:accuracy}
\vspace{-15pt}
\end{table}

\section{Inference, Control and Co-design}

The most attractive feature of our simulator is the existence of quickly computable gradients, which allows the use of much more efficient gradient-based optimization algorithms. In this section, we show the effectiveness of our differentiable simulator on gradient-based optimization tasks, including physical inference, control for soft robotics, and co-design of robotic arms. 

\subsection{Physical Parameter Inference} 
\begin{wrapfigure}{r}{.4\linewidth}
    \centering
    \vspace{-42pt}
    \label{fig:sysid}
    \vspace{-32pt}
\end{wrapfigure} 
\Name can be used to infer physical system properties given its observed motion, e.g. perform gradient descent to infer the relative densities of two colliding elastic balls (see figure above, ball A moving to the right hitting ball B, and ball B arrives the destination C). Gradient-based optimization infers that relative density of ball A is $2.26$, which constitutes to the correct momentum to push B to C. Such capability makes it useful for real-world robotic tasks such as system identification.

\subsection{Control}
We can optimize regression-based controllers for soft robots and efficiently discover stable gaits. The controller takes as input the state vector $\zz$, which includes target position, the center of mass position, and velocity of each composed soft component. In our examples, the actuation vector $\mathbf{a}$ for up to $16$ actuators is generated by the controller in each time step. During optimization, we perform gradient descent on variables $\WW$ and $\mathbf{b}$, where $\mathbf{a}=\tanh{(\WW\zz+\mathbf{b})}$ is the actuation-generating controller.

\begin{figure}[t]
    \centering
    \includegraphics[width=0.2\linewidth]{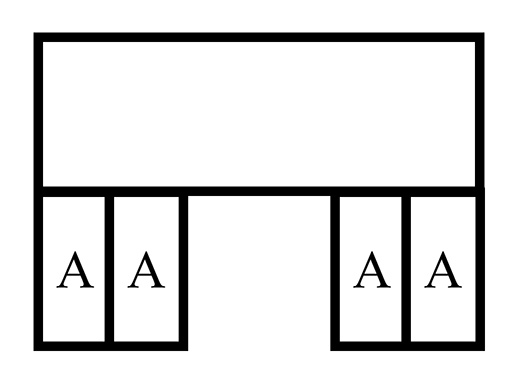}
    \includegraphics[height=0.15\linewidth]{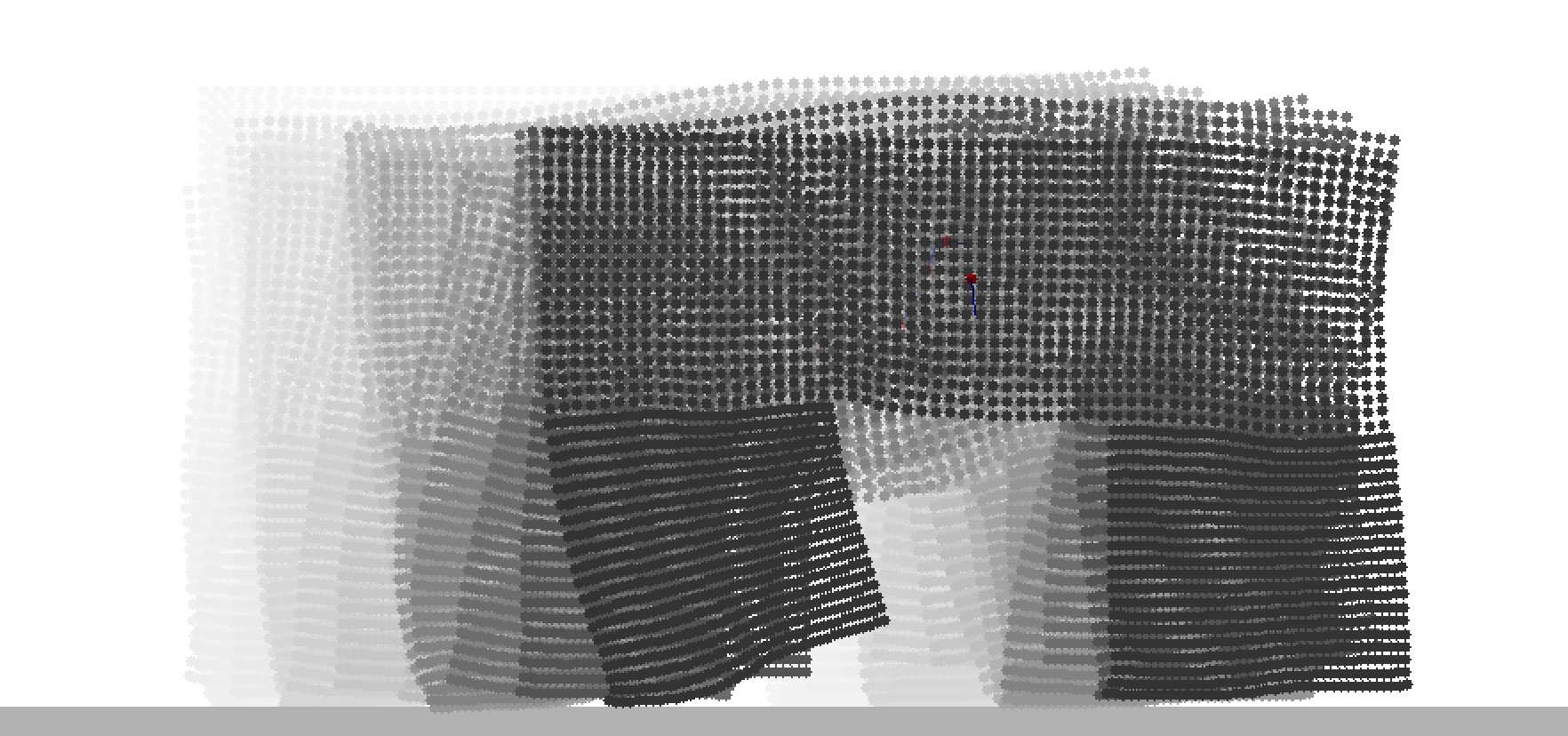}
    \includegraphics[height=0.15\linewidth]{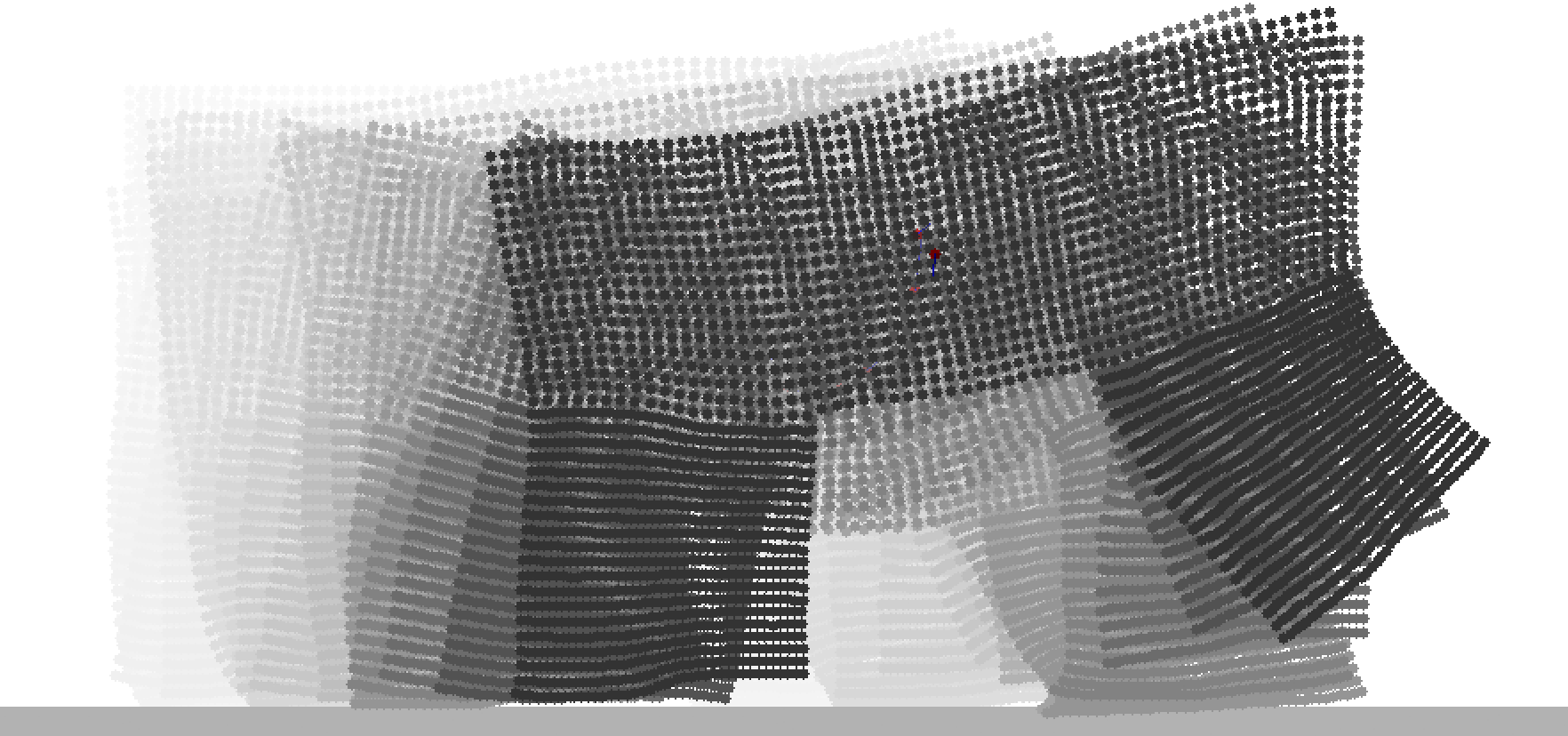}
    \caption{A soft 2D walker with controller optimized using gradient descent, aiming to achieve a maximum distance after $600$ simulation steps. The walker has four actuators (left, marked by letter `A's) with each capable of stretching or compressing in the vertical direction. The full walking animation (middle and right) is available in the  video.}
    \label{fig:runner2d}
    \vspace{-15pt}
\end{figure}

\begin{figure*}[t]
    \centering
    \includegraphics[width=0.95\linewidth]{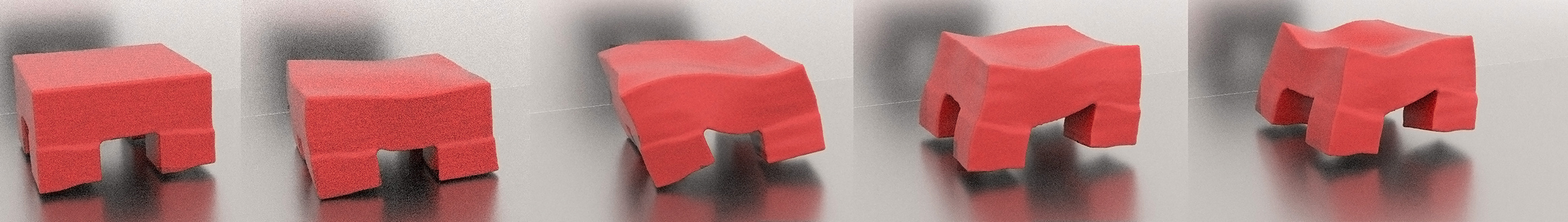}
    \caption{A 3D quadrupedal runner. Please see the supplemental video for more results.}
    \label{fig:3drobots}
    \vspace{-12pt}
\end{figure*}

We have designed a series of experiments including the 2D biped runner (\fig{fig:runner2d}) and robotic finger, and 3D quadrupedal runner (\fig{fig:3drobots}), crawler and robotic arm. Gradient-based optimizers successfully compute desired controllers within only tens or hundreds of iterations. Visual results are included in the supplemental video.

\begin{figure*}[t!]
    \centering
    \begin{subfigure}[t]{0.17\textwidth}
        \centering
        \includegraphics[height=1.2in]{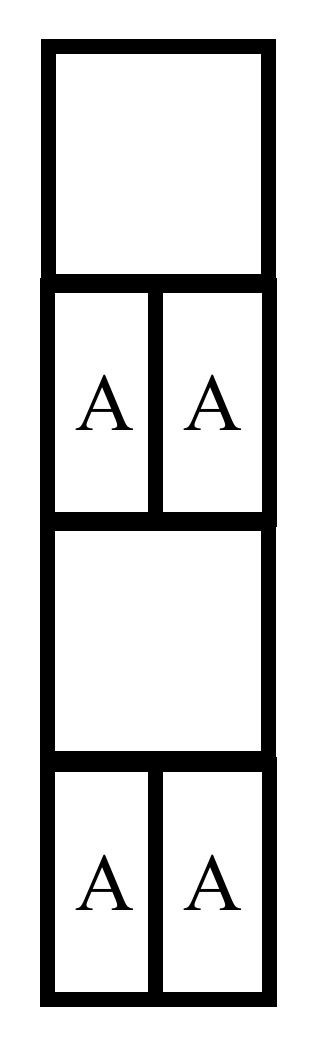}
        \caption{Actuation config}
    \end{subfigure}%
    ~
    \begin{subfigure}[t]{0.18\textwidth}
        \centering
        \includegraphics[height=1.2in]{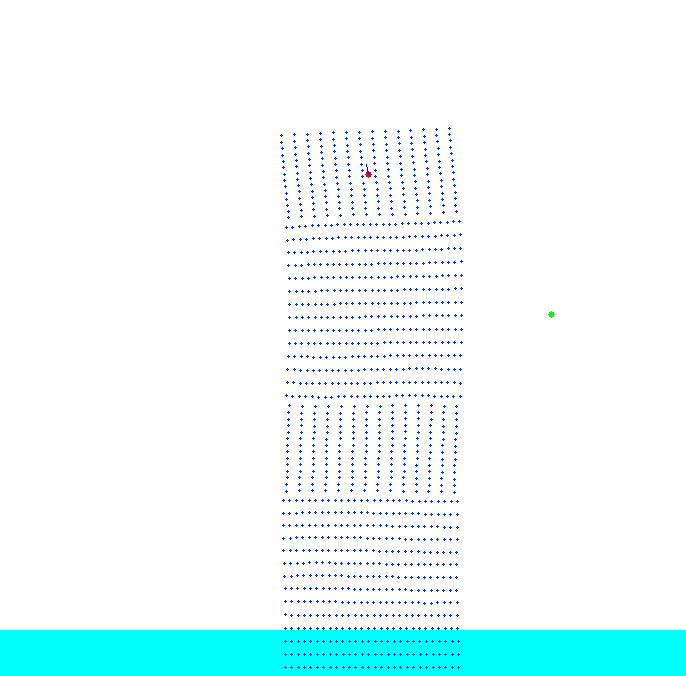}
        \caption{Resting pose}
    \end{subfigure}
    ~
    \begin{subfigure}[t]{0.18\textwidth}
        \centering
        \includegraphics[height=1.2in]{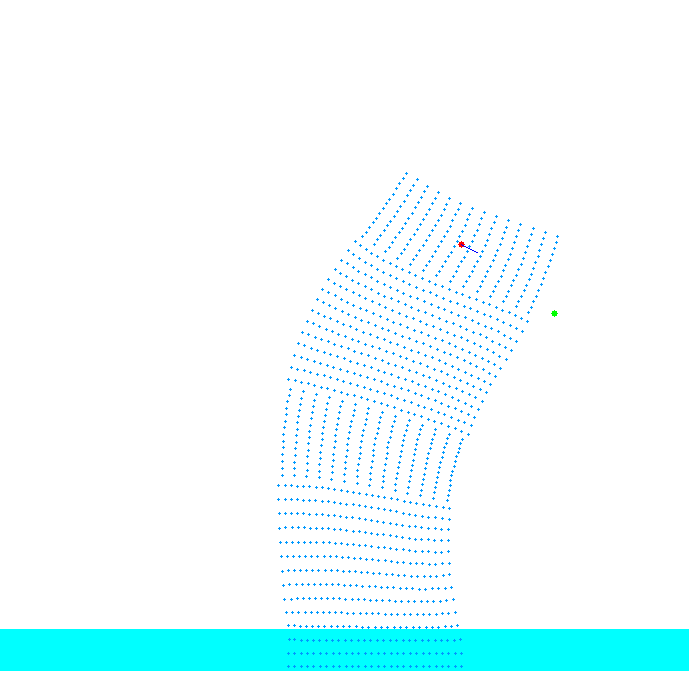}
        \caption{Final pose I}
    \end{subfigure}
    ~
    \begin{subfigure}[t]{0.18\textwidth}
        \centering
        \includegraphics[height=1.2in]{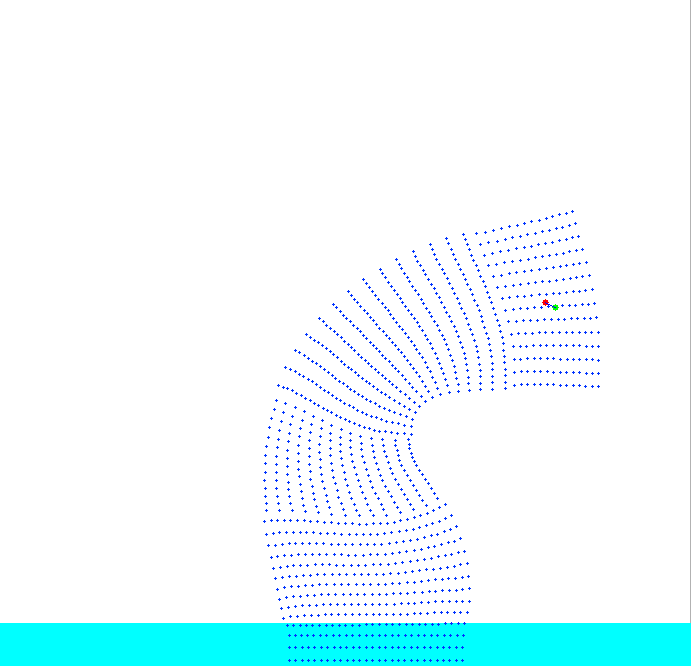}
        \caption{Final pose II}
    \end{subfigure}
    ~
    \begin{subfigure}[t]{0.18\textwidth}
        \centering
        \includegraphics[height=1.2in]{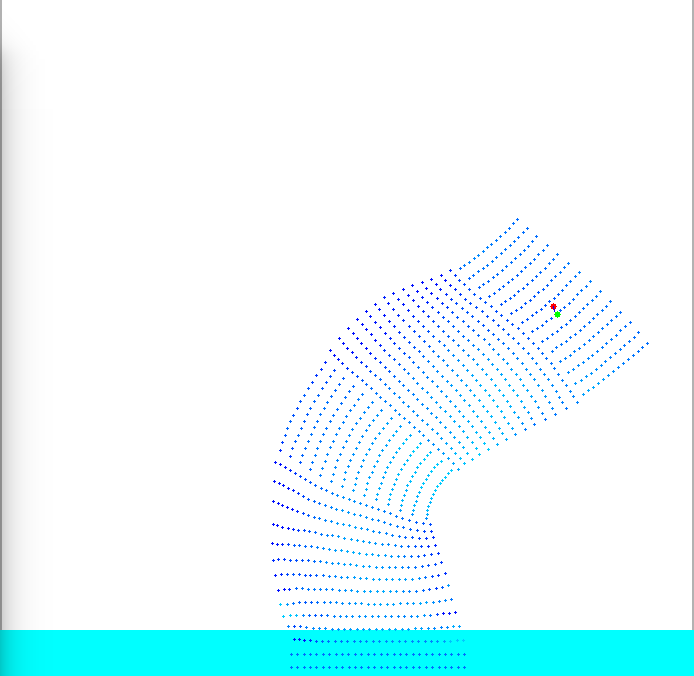}
        \caption{Final pose III} 
    \end{subfigure}%
    \caption{{\bf Final poses of the arm swing task.} Lighter colors refer to stiffer regions. (c) Final pose of the fixed-stiffness 300\% initial Young's modulus arm. (d) Final pose of the fixed-stiffness 300\% initial Young's modulus arm. (e) Final pose of the co-optimized arm. Actuation cost is 95.5\% that of the fixed 100\% initial Young's modulus arm and converges. Only the co-optimized arm is able to fully reach its target. The final optimized spatially varying stiffness of the arm has lower stiffness on the outside of the bend, and higher stiffness inside, promoting more bend to the left.  Qualitatively, this is similar in effect to the pleating on soft robot fingers.}
    \label{fig:arm}
    \vspace{-5pt}
\end{figure*}

\if(0)
\begin{figure}[t]
    \centering
    \includegraphics[width=0.48\linewidth]{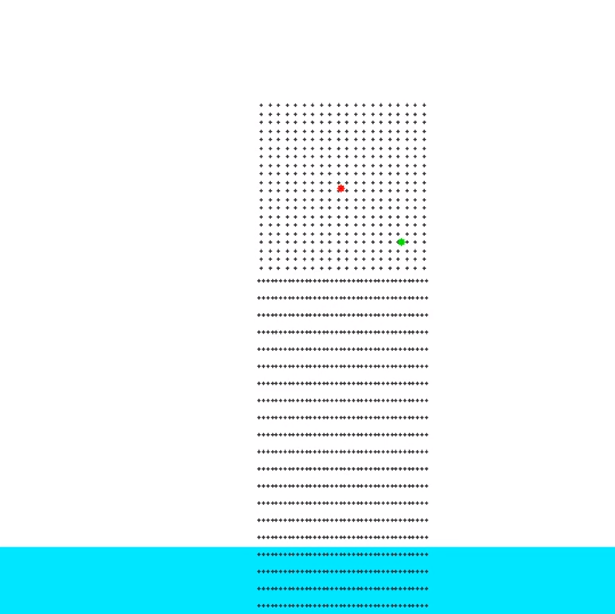}
    \includegraphics[width=0.48\linewidth]{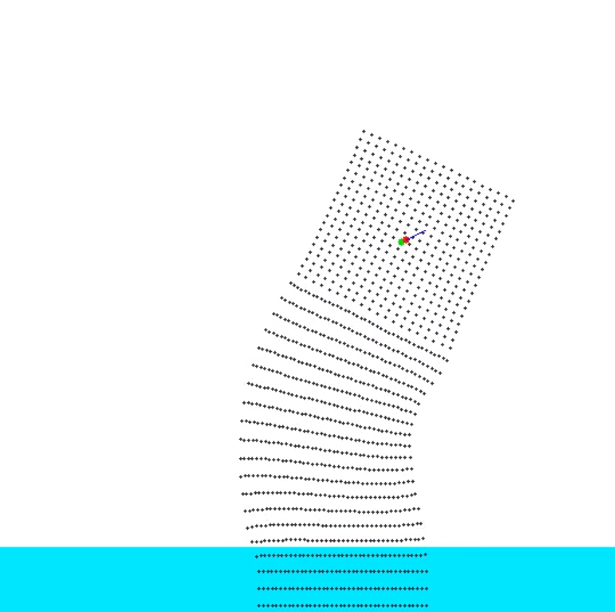}
    \caption{A soft finger trained to reach a target point. \yh{Placeholder. Needs better visualization/presentation here. Add quantitative results}.}
    \label{fig:finger}
    \vspace{-15pt}
\end{figure}
\fi

\begin{figure}[t]
    \centering
    \includegraphics[width=0.31\linewidth]{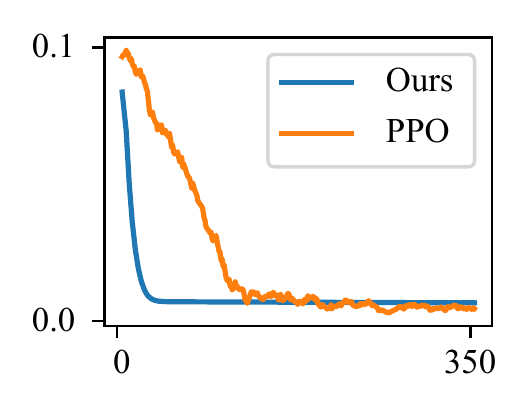}
    \includegraphics[width=0.325\linewidth]{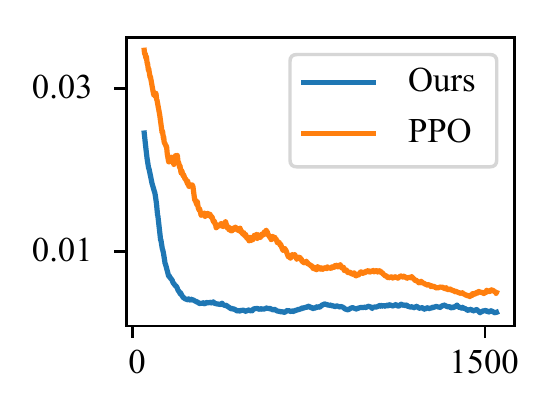}
    \includegraphics[width=0.325\linewidth]{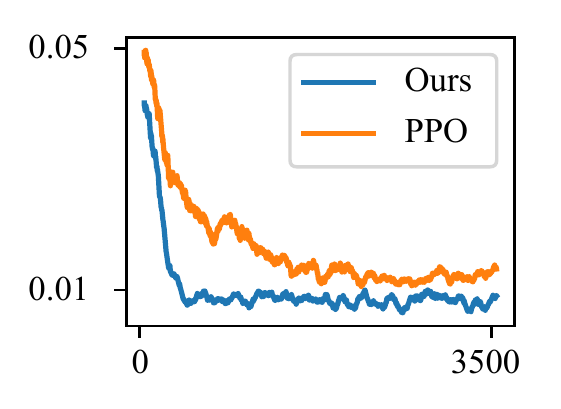}
    \vspace{-5pt}
    \caption{Gradient-free optimization using PPO and gradient-descent based on \Name, on the 2D finger task. Thanks to the accurate gradient information, even the most vanilla optimizer can beat state-of-the-art reinforcement learning algorithms by one order of magnitude regarding optimization speed. (Left) single, fixed target. (Middle) random targets. (Right) random targets, larger range. Curves are smoothed over 10, 100 and 100 iterations respectively. The $x$-axis is simulation iterations and $y$-axis the loss.}
    \label{fig:ppo}
    \vspace{-15pt}
\end{figure}

To emphasize the merits of gradient-based approaches, we compare our control method with proximal policy optimization (PPO)~\cite{Schulman2017Proximal}, a state-of-the-art reinforcement learning algorithm. PPO is an actor-critic method which relies on sampled gradients of a \emph{reward} function in order to optimize a policy. This sampling-based approach is model-free; it relies on gradients of the rewards with respect to controller parameters, but \emph{not} with respect to the physical model for updates. For our comparison, we use velocity projected onto the direction toward the goal as the reward.
\footnote{Note that this is functionally extremely similar to a distance loss; the cumulative reward $\int_t{=0}^T{v_{goal} dt} = D - \|x_T - x_{goal}\|$, where $D$ is the initial distance and $x_T$ and $x_{goal}$ represent world coordinates of the robot at time $T$ and of the goal, respectively.  As velocity toward the goal increases, final distance to the goal decreases.} We use a simplified single link version (with  only two adjacent actuators) of Fig.~\ref{fig:arm} and the 2D runner~\fig{fig:runner2d} as a benchmark. Quantitative results for the finger are shown in~\fig{fig:ppo}. We performed a similar comparison on the 2D walker, the controller optimized by \Name for the 2D walker starts functioning well within $20$ minutes; by comparison the policy recovered by PPO still chose nearly-random actions after over $4$ hours of training; demonstrating that for certain soft locomotion tasks our gradient-based method can be more efficient than model-free approaches.

\subsection{Co-design} 

\begin{figure}[t!]
    \centering
    \includegraphics[width=\linewidth]{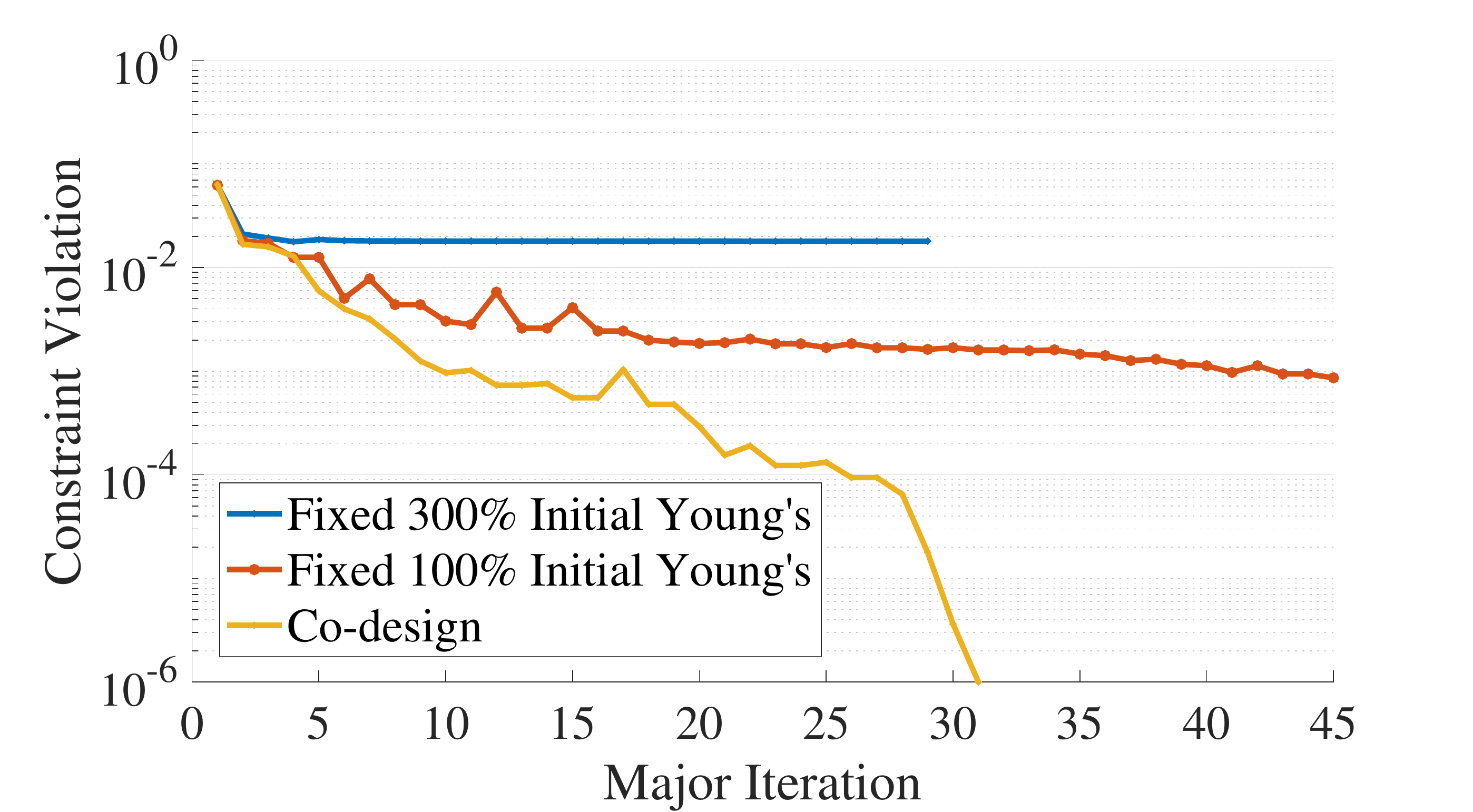}
    \vspace{-5pt}
    \caption{Convergence of the arm reaching task for co-design vs. fixed arm designs.  The fixed designs can make progress but not complete the task, while with co-design, the task can be completed and the actuation cost is lower. Constraint violation is the norm of two constraints: distance of end-effector to goal and mean squared velocity of the particles.}
    \label{fig:convergence}
    \vspace{-15pt}
\end{figure}

Our simulator is capable of not only providing gradients with respect to dynamics and controller parameters, but also with respect to structural design parameters, enabling co-design of soft robots. To demonstrate this, we designed a multi-link robot arm (two links, two joints each with two side-by-side actuators; all parts deformable).  Similar to shooting method trajectory optimization, actuation for each time step is solved for, along with the time-invariant Young's modulus of the system for each particle. In our task, we optimized the end-effector of the arm to reach a goal ball with final $0$ arm velocity, and minimized for actuation cost $\sum_{i=0}^N u_i^Tu_i dt$, where $u_i$ is the actuation vector at timestep $i$, and $N$ is the total number of timesteps. This is a \emph{dynamic} task and the target pose cannot be reached in a static equilibrium.  NLOPT's sequential least squares programming algorithm was used for optimization \cite{johnson2014nlopt}.  We
compared our co-design solution to fixed designs. The designed stiffness distribution is shown in \fig{fig:arm}, along with controls. The convergence for the different tasks can be seen in \fig{fig:convergence}. As can be seen, only the co-design arm fully converges to the target goal, and with lower actuation cost. Actuation for each chamber was clamped, and rnges of 30\%  to 400\% of a dimensionless initial Young's modulus were allowed and chosen large enough such as to require a swing instead of a simple bend.

\section{Discussion}

We have presented \Name, a differentiable simulator for soft robotics, and demonstrated how it can be deployed for inference, control, and co-design. \Name has the potential to accelerate the development of soft robots.
We have also developed a high-performance GPU implementation for \Name, which we plan to open source.

One interesting future direction is to couple our soft object simulation with rigid body simulation, as done in \cite{hu2018moving}. As derived in~\cite{fang2018async}, the $\Delta t$ limit for explicit time integration is $ C \Delta x \sqrt{\frac{\rho}{E}}$,
where $C$ is a constant close to one, $\rho$ is the density, and $E$ is the Young's modulus. That means for very stiff materials (\eg, rigid bodies), only a very restrictive $\Delta t$ can be used. However, a rigid body simulator should probably be employed in the realm of nearly-rigid objects and coupled with our deformable body simulator.
Combining our simulator with existing rigid-body simulators using Compatible Particle-in-Cell~\cite{hu2018moving} can be an interesting direction.

\section*{Acknowledgments}
We would like to thank Chenfanfu Jiang, Ming Gao and Kui Wu for the insightful discussions.
\newpage
\onecolumn
{\noindent\Large \textbf{Supplemental Document}}
\vspace{15pt}

In this document, we discuss the detailed steps for backward gradient computation in \Name, i.e. the differentiable Moving Least Squares Material Point Method (MLS-MPM)~\cite{hu2018moving}. Again, we summarize the notations in Table~\ref{table:notations}. We assume fixed particle mass $m_p$, volume $V_p^0$, hyperelastic constitutive model (with potential energy $\psi_p$ or Young's modulus $E_p$ and Poisson's ratio $\nu_p$) for simplicity. 

\begin{table}[h]
  \centering\small
  \setlength{\tabcolsep}{4pt}
\caption{List of notations for MLS-MPM.}
\scalebox{1.3}{
\begin{tabular}{cccl}
\toprule
Symbol & Type  & Affiliation & Meaning \\ 
\midrule
$\Delta t$ & scalar &   & time step size \\
$\Delta x$ & scalar &   & grid cell size \\
$\xx_p$ & vector & particle & position \\
$V_p^0$ & scalar  & particle & initial volume\\
$\vv_p$ & vector & particle & velocity \\
$\CC_p$ & matrix & particle & affine velocity field ~\cite{jiang:2015:apic} \\
$\PP_p$ & matrix & particle &  
PK1 stress ($\partial \psi_p/\partial \FF_p$)\\
$\sig_{pa}$ & matrix & particle &  
actuation Cauchy stress \\
$\AA_{p}$ & matrix & particle &  
actuation stress (material space) \\
$\FF_p$ & matrix & particle & deformation gradient\\
$\xx_i$ & vector & node  & position\\
$m_i$ & scalar  & node & mass\\
$\vv_i$ & vector & node & velocity\\
$\pp_i$ & vector & node & momentum, i.e. $m_i\vv_i$\\
$N$ & scalar &  & quadratic B-spline function\\
\bottomrule
\end{tabular}%
}
\label{table:notations}
\end{table}

\section{Variable dependencies}
The MLS-MPM time stepping is defined as follows:

\begin{eqnarray}
\ePpn \\
\emin \\
\eppin  \\
\evvin \\
\evvpnn \\
\eCCpnn  \\
\eFFpnn,  \\
\exxpnn \\
\end{eqnarray}

\newpage

The forward variable dependency is as follows:

\begin{eqnarray}
\xx_p^{n+1} & \dep & \xx_p^n, \vv_p^{n+1} \\
\vv_p^{n+1} & \dep & \xx_p^n, \vv_i^n \\
\CC_p^{n+1} & \dep & \xx_p^n, \vv_i^n \\
\FF_p^{n+1} & \dep & \FF_p^n, \CC_p^{n+1} \\
\pp_i^{n} & \dep & \xx_p^n, \CC_p^n, \vv_p^n, \PP_p^n, \FF_p^n \\
\vv_i^{n} & \dep & \pp_i^n, m_i^n \\
\PP_p^n & \dep & \FF_p^n, \sig^n_{pa} \\
m_i^{n} & \dep & \xx_p^n \\
\end{eqnarray}

During back-propagation, we have the following reversed variable dependency:

\begin{eqnarray}
\xx_p^{n+1}, \vv_p^{n+1}, \CC_p^{n+1}, \pp_i^{n+1}, m_i & \dep & \xx_p^n \\
\pp_i^{n} & \dep & \vv_p^n \\
\xx_p^{n+1} & \dep & \vv_p^{n+1} \\
\vv_p^{n+1}, \CC_p^{n+1} & \dep & \vv_i^{n} \\
\FF_p^{n+1}, \PP_p^n, \pp_i^n & \dep & \FF_p^n \\
\FF_p^{n+1} & \dep & \CC_p^{n+1} \\
\pp_i^{n} & \dep & \CC_p^n \\
\vv_i^{n} & \dep & \pp_i^n \\
\vv_i^{n} & \dep & m_i^n \\
\pp_i^{n} & \dep & \PP_p^n \\
\PP_p^n   & \dep & \sig_{pa}^n \\
\end{eqnarray}

We reverse swap two sides of the equations for easier differentiation derivation:

\begin{eqnarray}
\xx_p^n & \rdep & \xx_p^{n+1}, \vv_p^{n+1},  \CC_p^{n+1}, \pp_i^{n+1}, m_i\\
\vv_p^n & \rdep & \pp_p^{n}\\
\vv_p^{n+1} & \rdep & \xx_p^{n+1}\\
\vv_i^{n} & \rdep & \vv_p^{n+1}, \CC_p^{n+1}\\
\FF_p^n & \rdep & \FF_p^{n+1}, \PP_p^n, \pp_i^n \\
\CC_p^{n+1} & \rdep & \FF_p^{n+1}\\
\CC_p^n & \rdep & \pp_i^{n}\\
\pp_i^n & \rdep & \vv_i^{n}\\
m_i^n & \rdep & \vv_i^{n}\\
\PP_p^n & \rdep & \pp_i^{n}\\
\sig_{pa}^n & \rdep & \PP_p^n \\
\end{eqnarray}

\newcommand{\dN}{\pd{N(\xx_i-\xx_p^n)}{\xx_i}}
\newcommand{\dNa}{\pd{N(\xx_i-\xx_p^n)}{\xx_{i\alpha}}}
\newcommand{\N}{N(\xx_i-\xx_p^n)}
In the following sections, we derive detailed gradient relationships, in the order of actual gradient computation. The frictional boundary condition gradients are postponed to the end since it is less central, though during computation it belongs to grid operations. 
Back-propagation in \Name is essentially a reversed process of forward simulation. The computation has three steps, backward particle to grid (P2G), backward grid operations, and backward grid to particle (G2P).

\newpage
\section{Backward Particle to Grid (P2G)}

%%%%%%%%%%%%%%%%%%%%%%%%%%%%%%%%%%%%%%%%%%%%%%%%%%%%%%%%%%%%%
(A, P2G) For $\vv_p^{n+1}$, we have 

\begin{eqnarray}
\exxpnn\\
    \Longrightarrow \D{\vv_{p\alpha}^{n+1}} &=& \left[\chain{\xx_p^{n+1}}{\vv_p^{n+1}}\right]_{\alpha} \\
    &=& \Delta t \D{\xx_{p\alpha}^{n+1}}.
\end{eqnarray}
%%%%%%%%%%%%%%%%%%%%%%%%%%%%%%%%%%%%%%%%%%%%%%%%%%%%%%%%%%%%%

%%%%%%%%%%%%%%%%%%%%%%%%%%%%%%%%%%%%%%%%%%%%%%%%%%%%%%%%%%%%%
(B, P2G) For $\CC_p^{n+1}$, we have
\begin{eqnarray}
\eFFpnn \\
    \Longrightarrow \D{\CC_{p\alpha\beta}^{n+1}} &=& \left[\chain{\FF_p^{n+1}}{\CC_p^{n+1}}\right]_{\alpha\beta} \\
    &=& \Delta t \sum_\gamma\D{\FF_{p\alpha\gamma}^{n+1}}\FF_{p\beta\gamma}^n.
\end{eqnarray}

Note, the above two gradients should also include the contributions of $\D{\vv_p^{n}}$ and $\D{\CC_p^{n}}$ respectively, with $n$ being the next time step.

%%%%%%%%%%%%%%%%%%%%%%%%%%%%%%%%%%%%%%%%%%%%%%%%%%%%%%%%%%%%%

%%%%%%%%%%%%%%%%%%%%%%%%%%%%%%%%%%%%%%%%%%%%%%%%%%%%%%%%%%%%%
(C, P2G) For $\vv_i^n$, we have
\begin{eqnarray}
    \evvpnn\\
    \eCCpnn\\
    \Longrightarrow \D{\vv_{i\alpha}^n} &=& \left[\sum_p \chain{\vv_p^{n+1}}{\vv_i^n} + \sum_p \chain{\CC_p^{n+1}}{\vv_i^n}\right]_{\alpha} \\
    &=& \sum_p\left[\D{\vv_{p\alpha}^{n+1}}\N +\Dinv\N\sum_\beta \D{\CC_{p\alpha\beta}^{n+1}} (\xx_{i\beta}-\xx_{p\beta})\right].
\end{eqnarray}
%%%%%%%%%%%%%%%%%%%%%%%%%%%%%%%%%%%%%%%%%%%%%%%%%%%%%%%%%%%%%

\section{Backward Grid Operations}

%%%%%%%%%%%%%%%%%%%%%%%%%%%%%%%%%%%%%%%%%%%%%%%%%%%%%%%%%%%%%
(D, grid) For $\pp_i^n$, we have
\begin{eqnarray}
\evvin\\
    \Longrightarrow \D{\pp_{i\alpha}^n} &=& \left[\chain{\vv_i^{n}}{\pp_i^n}\right]_{\alpha}\\
    &=& \D{\vv_{i\alpha}^{n}} \frac{1}{m_i^n}.
\end{eqnarray}
%%%%%%%%%%%%%%%%%%%%%%%%%%%%%%%%%%%%%%%%%%%%%%%%%%%%%%%%%%%%%

%%%%%%%%%%%%%%%%%%%%%%%%%%%%%%%%%%%%%%%%%%%%%%%%%%%%%%%%%%%%%
(E, grid) For $m_i^n$, we have
\begin{eqnarray}
\evvin\\
    \Longrightarrow \D{m_{i}^n} &=& \chain{\vv_i^{n}}{m_i^n}\\
    &=& -\frac{1}{(m_i^n)^2}\sum_\alpha \pp_{i\alpha}^{n}\D{\vv_{i\alpha}^{n}} \\
    &=& -\frac{1}{m_i^n}\sum_\alpha \vv_{i\alpha}^{n}\D{\vv_{i\alpha}^{n}}.
\end{eqnarray}
%%%%%%%%%%%%%%%%%%%%%%%%%%%%%%%%%%%%%%%%%%%%%%%%%%%%%%%%%%%%%

\section{Backward Grid to Particle (G2P)}
%%%%%%%%%%%%%%%%%%%%%%%%%%%%%%%%%%%%%%%%%%%%%%%%%%%%%%%%%%%%%
(F, G2P) For $\vv_p^n$, we have
\begin{eqnarray}
\eppin \\
\Longrightarrow \D{\vv_{p\alpha}^n} &=& \left[\sum_i \chain{\pp_p^{n}}{\vv_p^n}\right]_{\alpha} \\
    &=& \sum_i \N m_p \D{\pp_{i\alpha}^n}.
\end{eqnarray}
%%%%%%%%%%%%%%%%%%%%%%%%%%%%%%%%%%%%%%%%%%%%%%%%%%%%%%%%%%%%%

%%%%%%%%%%%%%%%%%%%%%%%%%%%%%%%%%%%%%%%%%%%%%%%%%%%%%%%%%%%%%
(G, G2P) For $\PP_p^n$, we have
\begin{eqnarray}
\eppin\\
    \Longrightarrow \D{\PP_{p\alpha\beta}^n} &=& \left[\chain{\pp_i^{n}}{\PP_p^n}\right]_{\alpha\beta}\\
    &=& -\sum_i\N \frac{4}{\Delta x^2}\Delta t V_p^0 \sum_\gamma \D{\pp_{i\alpha}^n}   \FF_{p_\gamma\beta}^{n}(\xx_{i\gamma}-\xx_{p\gamma}^n).
\end{eqnarray}
%%%%%%%%%%%%%%%%%%%%%%%%%%%%%%%%%%%%%%%%%%%%%%%%%%%%%%%%%%%%%

%%%%%%%%%%%%%%%%%%%%%%%%%%%%%%%%%%%%%%%%%%%%%%%%%%%%%%%%%%%%%
(H, G2P) For $\FF_p^{n}$, we have
\begin{eqnarray}
\eFFpnn \\
\ePpn \\
\eppin \\
    \Longrightarrow \D{\FF_{p\alpha\beta}^n} &=& \left[\chain{\FF_p^{n+1}}{\FF_p^n} + \chain{\PP_p^n}{\FF_p^n}+\chain{\pp_i^n}{\FF_p^n}\right]_{\alpha\beta} \\
    &=& \sum_\gamma\D{\FF_{p\gamma\beta}^{n+1}}(\II_{\gamma\alpha}+\Delta t \CC_{p\gamma\alpha}^{n+1}) + \sum_\gamma \sum_\eta \D{\PP_{p\gamma\eta}} \frac{\partial ^2\Psi_p}{\partial \FF^n_{p\gamma\eta} \partial \FF_{p\alpha\beta}^n}+\sum_\gamma\D{\PP_{p\alpha\gamma}^{n}}\sig_{pa\beta\gamma}\\
    &&+\sum_i -\N \sum_\gamma \D{\pp_{i\gamma}^n} \Dinv \dt V_p^0\PP_{p\gamma\beta}^n (\xx_{i\alpha} - \xx_{p\alpha}^n). % Note: there is a transpose here
\end{eqnarray}
%%%%%%%%%%%%%%%%%%%%%%%%%%%%%%%%%%%%%%%%%%%%%%%%%%%%%%%%%%%%%

%%%%%%%%%%%%%%%%%%%%%%%%%%%%%%%%%%%%%%%%%%%%%%%%%%%%%%%%%%%%%
(I, G2P) For $\CC_p^n$, we have
\begin{eqnarray}
\eppin\\
    \Longrightarrow \D{\CC_{p\alpha\beta}^n} &=& \left[\sum_i\chain{\pp_i^{n}}{\CC_p^n}\right]_{\alpha\beta}\\
    &=& \sum_i \N \D{\pp_{i\alpha}^{n}} m_p (\xx_{i\beta}-\xx_{p\beta}^n).
\end{eqnarray}
%%%%%%%%%%%%%%%%%%%%%%%%%%%%%%%%%%%%%%%%%%%%%%%%%%%%%%%%%%%%%

%%%%%%%%%%%%%%%%%%%%%%%%%%%%%%%%%%%%%%%%%%%%%%%%%%%%%%%%%%%%%
(J, G2P) For $\xx_p^n$, we have
\begin{eqnarray}
    \exxpnn \\
    \evvpnn \\
    \eCCpnn \\
    \eppin\\
    \emin\\
    \GG_p &:=& \left(-\frac{4}{\Delta x^2} V_p^0\dt\PP_p^n\FF_p^{nT} + m_p \CC_p^n\right)\\
    \Longrightarrow&&\\ \D{\xx_{p\alpha}^{n}} &=& \left[\chain{\xx_p^{n+1}}{\xx_p^n} +\chain{\vv_p^{n+1}}{\xx_p^n} +\chain{\CC_p^{n+1}}{\xx_p^n} +\chain{\pp_i^{n}}{\xx_p^n}+\chain{m_i^{n}}{\xx_p^n}\right]_{\alpha} \\
    &=&\D{\xx_{p\alpha}^{n+1}}\\
    &&+\sum_i \sum_\beta \D{\vv_{p\beta}^{n+1}}\dNa \vv_{i\beta}^n\\
    &&+\sum_i \sum_\beta \Dinv \left\{-\D{\CC_{p\beta\alpha}^{n+1}}\N\vv_{i\beta} + \sum_\gamma \D{\CC_{p\beta\gamma}^{n+1}}\dNa\vv_{i\beta}(\xx_{i\gamma}-\xx_{p\gamma})\right\}\\
    &&+\sum_i \sum_\beta \D{\pp_{i\beta}^{n}}\left[\dNa \left(m_p \vv_{p\beta}^n + [\GG_{p}(\xx_i-\xx_p^n)]_{\beta}\right)-\N\GG_{p\beta\alpha} \right]\\
    &&+m_p \sum_i \D{m_i^n} \dNa\\
\end{eqnarray}

(K, G2P) For $\sig_{pa}^{n}$, we have
\begin{eqnarray}
\ePpn \\
    \Longrightarrow \D{\sig_{pa\alpha\beta}^{n}} &=& \left[\chain{\PP_p^{n}}{\sig_{p\alpha}^n}\right]_{\alpha\beta} \\
    &=& \sum_\gamma\D{\PP_{p\gamma\beta}^{n+1}}\FF_{p\gamma\alpha}^{n}. % Note: there is a transpose here
\end{eqnarray}

\newpage

\section{Friction Projection Gradients}
When there are boundary conditions:

(L, grid) For $\vv_i^n$, we have
\begin{eqnarray}
    l_{i\nn} &=& \sum_\alpha \vv_{i\alpha} \nn_{i\alpha}\\
    \vv_{i\tt}&=&\vv_i-l_{i\nn}\nn_i\\
    l_{i\tt} &=& \sqrt{\sum_\alpha \vv_{i\tt\alpha}^2 +\varepsilon } \\
    \hat{\vv}_{i\tt}&=&\frac{1}{l_{i\tt}}\vv_{i\tt}\\
     l_{i\tt}^*&=&\max\{l_{i\tt}+c_i\min\{l_{i\nn}, 0\}, 0\}\\
     \vv_i^*&=&l^*_{i\tt}\hat{\vv}_{i\tt}+\max\{l_{i\nn}, 0\}\nn_i\\
     H(x)&:=&[x\ge 0]\\
     R&:=&l_{i\tt}+c_i\min\{l_{i\nn,0}\}\\
    \Longrightarrow  \D{l_{i\tt}^*}&=&\sum_\alpha \D{\vv^*_{i\alpha}} \hat{\vv}_{i\tt\alpha}\\
    \D{\hat{\vv}_{i\tt}} &=& \D{\vv^*_{i\alpha}}l_{i\tt}^*\\
    \D{l_{i\tt}}&=&-\frac{1}{l^{2}_{i\tt}}\sum_\alpha \vv_{i\tt\alpha} \D{\hat{\vv}_{i\tt\alpha}}+\D{l_{i\tt}^*}H(R)\\
    \D{\vv_{i\tt\alpha}}&=&\frac{\vv_{i\tt\alpha}}{l_{i\tt}}\D{l_{i\tt}}+\frac{1}{l_{i\tt}}\D{\hat{\vv}_{i\tt\alpha}}\\
    &=&\frac{1}{l_{i\tt}}\left[\D{l_{i\tt}}\vv_{i\tt\alpha}+\D{\hat{\vv}_{i\tt\alpha}}\right]\\
    \D{l_{i\nn}} &=& -\left[\sum_\alpha \D{\vv_{i\tt\alpha}}\nn_{i\alpha}\right]+\D{l_{i\tt}^*}H(R) c_i H(-l_{i\nn})+\sum_\alpha H(l_{i\nn})\nn_{i\alpha}\D{\vv_{i\alpha}^*}\\
    \D{\vv_{i\alpha}} &=& \D{l_{i\nn}}\nn_{i\alpha}+\D{\vv_{i\tt\alpha}}\\
\end{eqnarray}

%%%%%%%%%%%%%%%%%%%%%%%%%%%%%%%%%%%%%%%%%%%%%%%%%%%%%%%%%%%%%j

%%%%%%%%%%%%%%%%%%%%%%%%%%%%%%%%%%%%%%%%%%%%%%%%%%%%%%%%%%%%%%%%%%%%%%%%%%%%%%%%

\clearpage %force bib to end
\bibliographystyle{IEEEtran}
\bibliography{diffphys}

\end{document}